# Tensor-based approach to accelerate deformable part models.

D. V. Parkhomenko,  I. L. Mazurenko

Mathematical Theory of Intelligent Systems of Moscow State University

## Abstract

*This article provides next step towards solving speed bottleneck of any system that intensively uses convolutions operations (e.g. CNN). Method described in the article is applied on deformable part models (DPM) algorithm. Method described here is based on multidimensional tensors and provides efficient tradeoff between DPM performance and accuracy. Experiments on various databases, including Pascal VOC, show that the proposed method allows decreasing a number of convolutions up to 4.5 times compared with DPM v.5, while maintaining similar accuracy. If insignificant accuracy degradation is allowable, higher computational gain can be achieved. The method consists of filters tensor decomposition and convolutions shortening using the decomposed filter. Mathematical overview of the proposed method as well as simulation results are provided.*

## 1. Introduction

Original Deformable Part Models (DPM) were presented as a method for object detection in 2010 [1]. They were proposed for PASCAL VOC [2] challenge. The results shown by DPM approach at the competition outperformed nearest competitors in terms of detection accuracy versus false detection rate. Further, various modernizations of DPM were proposed. For reference, we choose DPM v.5. Lately, the authors investigated different other models for visual object detection, among them bag of words, convolutional neural networks (CNN) etc. Even though such new approaches occur, DPM method still shows nearly leading performance and has certain advantages in case of large object appearance variations, like in case of person detection and human pose estimation applications [3]. That is why DPM remains one of top models in object detection domain.

Main drawback of the DPM method in real world object detection is its computational complexity. Execution time is a bottleneck, as the method comprises consequent convolutions operations and number of such convolutions linearly depends on a number of pixels in the processed image.

Recent works propose a number of techniques to handle this problem. Among them we can point to FFT-based approach [4], cascade approach [5,8], coarse-to fine approach [6], various look-up tables and locality sensitive hashes techniques [7] and filter singular value decomposition (SVD) [8]. Authors stated that these methods can significantly reduce detection cost up of deformable part models up to an order of magnitude. Nevertheless, further acceleration is pivotal for real-time applications, as DPM still remains in non-real time domain.

We propose an acceleration technique which is compatible with some other types of speeding up the DPM algorithm. As experiments show, combining different approaches with the proposed one improves performance of many of them. The proposed tensor-based approach does not require time consuming Support Vector Machines (SVM) retraining and can be used in with already pre-trained models. Natural tuned-up tradeoff between accuracy and speed of calculations can be viewed as additional advantage of the proposed method, especially valuable for real-time applications.

The rest of the paper is organized as follows. Section 2 shortly reviews a conventional DPM-based object detection approach. Section 3 considers the related works. Introduction of the proposed tensor-based approach is given in Section 4. Experiments specification, comparisons and test results are presented in Section 5. Section 6 concludes the paper.

## 2. DPM overview

This Part gives a brief review of DPM v.5. Discriminatively trained deformable part models [1, 9] can be considered as a learning-based system for detecting and localizing objects in images. For detection process DPM uses mixtures of models, each composed of a number of filters and predefined dependencies between them.

Each DPM model consists of a root filter $f_0$, $n$ part filters $f_i$ and $n$ deformations cost coefficients $d_i$, where $i=1,...,n, n>0$. Each filter is a matrix of features, some variation of HOG features are commonly used here. Sizes of various filters typically differ. The root and the parts are usually connected in a star-like structure, and offsets from the original positions are set by the deformation terms. High level description of DPM-based detection procedure can be found below.

Let object hypothesis $\Im=\{p_0,p_1,...,p_n\}$, where $p_0$ is a root filter location and each $p_i$, $i>0$ is a location of the $i$-th part filter. Than a detection of the object given by hypothesis $\Im$ is performed by calculation of a score $s(\Im)$ in the position $\{p_0,p_1,...,p_n\}$ and further comparison of $s(\Im)$ with a predefined threshold. The detection score $s(\Im)$ is computed using the formula:

$$s(\Im) = f_0\varphi(p_0) + \sum_{i=1}^{n}(f_i \cdot \varphi(p_i) - d_i\psi(p_i, p_0)),$$

where $\varphi(p)$ defines HOG feature vector at position $p$, and $\psi()$ is a bilinear deformation form. To compute the best part location, the best part position score is substituted by the deformation cost with displacements followed by the *argmax* operation:

$$p_i = \arg\max_{p}(f_i\varphi(p) - d_i\psi(p, p_0)),$$

where the position $p$ runs over all possible part positions.

Besides score computing part, there is a features extraction block in the DPM-based detection approach. But as can be seen from [5, 6], most time is spent for calculation of convolutions and part positions detection in the score estimation block. From our experiments we found that score calculation is 20 times more time consuming operation than the feature extraction in average, which is presented in Section 5. Section 6 concludes the paper.

## 3. Related work

First significant improvement was proposed by DPM authors [5]. So called Cascades approach allows omitting calculation of convolutions in knowingly futile points on the one hand, and does not perform filter convolutions in the points with the score which is already high enough. Mathematically, cascade approach can be described by converting star-structured DPM to the cascade-based structure.

One can see that DPM with cascades in the case of the worst input image does not provide a speed-up gain. But in real life applications objects of interest are concentrated in some specific area (see Figure 1 for illustration) and do not cover the whole image. Because of that, in practice efficient pruning of unpromising hypothesis can accelerate an overall detection time by an order of magnitude.

The authors of [6] have shown that a model with low resolution can prune a lot of hypothesis with a low cost, yielding up to ten-fold speed-up. This multiple resolution hierarchical based method is known as coarse-to-fine approach and is complementary to cascades.

The authors of [8] have made an observation that a root filter, considered as a set of 2D feature maps, can be constructed to be low-rank in terms of linear dependencies between map elements. To be more precise, each of 2D levels of the root is a low rank matrix. Construction of such low rank root requires updating the learning procedure: the authors of [8] proposed a modification of a proximal gradient algorithm to make the root rank low. Exploiting low rank feature of the root allows performing fewer multiplications. Another idea is to exploit a spatial redundancy of the image, incorporating the so called Neighborhood aware cascades. In couple with look-up tables (LUT) for histogram of oriented gradients these techniques can speed-up the reference DPM by up to 4 times.

An alternative method was proposed in [4]. As a convolution in spectral domain can be represented as a summation, one can decrease a number of expensive multiplications by applying an FFT transform.

Next bunch of methods is devoted to accelerating DPM for detecting multiple objects. Locality sensitive hashes are widely used in high dimensional space problems. Authors of [10] have proposed their own binary descriptor called the Winner-Take-All hash (WTA) to replace a linear convolution with an ordinal convolution by using WTA. Their empirical tests show that DPM can supply a significant number of objects for detection with some degradation of accuracy. As stated in [7] LSH coupled with LUT of HOGs and a pyramid of templates can handle 20 PASCAL object categories at a frame rate of 30 fps using CPU with 6 cores and 32GB of RAM.

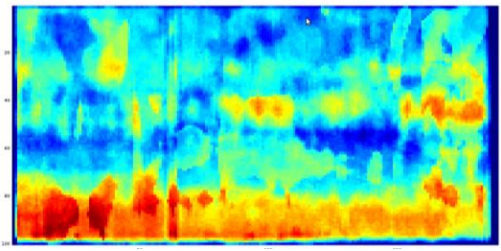

Figure 1: Person position probability map for one size images in representative image base. Intensity of pixel devoted to person detection probability in this position.

Sparselets [11,12] can be considered as a basis in the linear space of filters. Such sparselet functions were widely adopted for multi category DPM acceleration.

To summarize it, there are many methods to decrease a computational complexity in a single and a multi class DPM detection scenarios. Here we present a novel approach, which is different from the existing ones. Its worth to say that the proposed approach can be combined with some of them to increase a calculation speed in a very significant way. Among advantages of the proposed method we can underline conservative memory requirements for storage of the models (up to an order of magnitude in comparison with the conventional DPM).

## 4. Tensor approach

Without loss of generality, hereinafter we will consider tensors [13] of order 3. Each tensor of order 3 can be represented as a 3-dimensional matrix $T = \{t_{ijk}\}$, $0<i<N$, $0<j<M$, $0<k<L$. Let us denote with $\otimes$ a tensor product operation, i.e. for two tensors $C=\{c_1,...,c_N\}$, $D=\{d_1,...,d_M\}$ of order 1:

$$C \otimes D = \begin{pmatrix} c_1 d_1 & ... & c_1 d_M \\ & ... & \\ c_N d_1 & ... & c_N d_M \end{pmatrix}.$$

Tensor summation is performed in the same way as matrix addition.

Each tensor has a canonic polyadic (CP) decomposition [14]. For the given positive integer rank R, CP is a representation of the tensor as a sum of order one tensor products, see Figure 2.

$$T = a_1 \otimes b_1 \otimes c_1 + ... + a_R \otimes b_R \otimes c_R$$

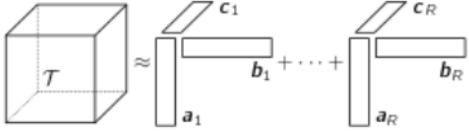

Figure 2: Rank "R" Canonic Polyadic tensor "T" decomposition.

CP decomposition choice is derived from its optimality in the Frobenius norm domain:

$$\|T\|_F = \sqrt{\sum_{i,j,k} t_{ijk}^2}\;,\;\; \|T\text{-}CP(T)\|_F \to min$$

Decomposition calculation can be done using the iterative alternating least squares (ALS) method [15]. Consider a function $f(A,B,C) = \| T\text{-}sum(a_i \otimes b_i \otimes c_i) \|_F$, where $a_i$, $b_i$, $c_i$ are columns of matrices $A,B,C$ respectively. ALS seeks for a local minimum $f(A,B,C) = min$ using the following iteration of the processing cycle:

$$\mathbf{A}^{(n+1)} = \underset{\mathbf{A} \in \mathbb{R}^{n_1 \times r}}{\operatorname{argmin}} f(\mathbf{A}, \mathbf{B}^{(n)}, \mathbf{C}^{(n)}),$$
$$\mathbf{B}^{(n+1)} = \underset{\mathbf{B} \in \mathbb{R}^{n_2 \times r}}{\operatorname{argmin}} f(\mathbf{A}^{(n+1)}, \mathbf{B}, \mathbf{C}^{(n)}),$$
$$\mathbf{C}^{(n+1)} = \underset{\mathbf{C} \in \mathbb{R}^{n_3 \times r}}{\operatorname{argmin}} f(\mathbf{A}^{(n+1)}, \mathbf{B}^{(n+1)}, \mathbf{C}).$$

Each micro iteration step is considered as the least squares problem. Convergence is achieved after the iterations limit is reached or after the Frobenius norm $f(A,B,C)$ becomes less than some predefined $\varepsilon>0$.

Kruskal in [16] found the CP decomposition uniqueness sufficient condition:

**Theorem** (Kruskal):
Let $T = a_1 \otimes b_1 \otimes c_1 + ... + a_R \otimes b_R \otimes c_R$ and $S_A=\{[a_i]\}$, $S_B=\{[b_i]\}$, $S_C=\{[c_i]\}$. If
$$R \leq 0.5(k(S_A) + k(S_B) + k(S_C)) - 1,$$
*where $k(S)$ is maximum number r, such all subsets of r column vectors of S are linearly independent. Then T has rank R and its expression as a rank R tensor is essentially unique.*

Decomposition is unique for the vast majority of ranks and tensor sizes. In our experiments, all tuples (rank, tensor sizes) that can be met in practice lead to an unambiguous CP form.

In this paper, different methods are used to accelerate the DPM score estimation procedure. First, for each filter of the mixture we search for the integer value $R>0$ and perform rank $R$ CP decomposition of this filter, considering it as a tensor. This extracted "sum of products form" allows faster calculation of convolutions. Second, for each filter we found a set of thresholds: $positive=\{t_1,...,t_R\}$. Usage of these thresholds allows convolutions shortening i.e. pruning unpromising convolutions. Despite the fact that the new $R$ integers need to be stored in memory for every filter, an overall required volume decreases, as CP form occupies less space than keeping the filter as a set of matrices.

Exploiting a filter as a tensor tends to extract its inner dependencies in 3D domain. A fact that dependencies among all tensor dimensions are more preferable than dependencies among 2 directions was verified by experiments.

The decomposed structure of the filter admits efficient convolution calculation. For image size $N'$, $M'$ and HOG feature length $L$ a standard score calculation method for one filter requires $\alpha = N'\cdot M'\cdot N\cdot M\cdot L$ multiplications. Using decomposed filter we can calculate convolution among x-direction first for all image points, then convolve resulting 2D convolution map with y-direction decomposition part, and then a correlation with z-axes decomposition vector is applied to the result. This is done for every rank 1 tensor in the decomposition. This procedure is repeated $R$ times, and the results are summed up:

$$I \circ T = I \circ \sum_{r=1}^{R} a_r \circ b_r \circ c_r = I \circ \sum_{r=1}^{R}\sum_{n=1}^{N}\sum_{m=1}^{M}\sum_{l=1}^{L} a_n(r)b_m(r)\cdot c_k(r)$$

$$I \circ T = \sum_{r=1}^{R}((I \circ a_r) \circ b_r) \circ c_r\;,$$

where $\circ$ denotes a correlation operator. This procedure requires $R\cdot(N'M'\cdot L + N'\cdot M'\cdot M + N'\cdot M'\cdot N) = R\cdot N'\cdot M'\cdot(L+M+N)$ multiplications. A combination of 1D correlations requires fewer multiplications if the rank R is small. It can be observed that both speed-up of calculation and a memory reduction gain (required space to store a filter as 3D matrix relative to the space to store CP decomposed tensor) is given by the formula:

$$\frac{N \cdot M \cdot L}{R \cdot (N + M + L)} \quad (1)$$

A rank R detection can be done by searching for a such minimal $r$ that:

$$\|f - CP(f,r)\|_F < e \cdot \left(\frac{r \cdot (N+M+L)}{N \cdot M \cdot L}\right)^2,$$

where *CP(f,r)* stands for a filter *"f"* tensor decomposition of rank *r*. For some specific filters a search for the finer rank may be required. For simplicity, we did different tests with various decomposition ranks and estimated an overall accuracy of detection. The first step of the best rank binary search was being given by the formula provided above.

After a rank is determined and CP decomposition is done, we determined positive thresholds by extracting positive examples from the train image set, succeeded by estimation of all $t_i$, $i=1,...,R$ for every filter $f$:

$$t_i = \min_{\phi \in positives}(\phi \circ CP(f,R)[i]),$$

where set *positives* is a set of positive detections extracted from the training set and *CP(f,R)[i]* is a rank *R* canonic polyadic decomposition of the filter abridged to the first *"i"* summands.

These are the tightest thresholds that make no mistakes on positive examples set.

Below we provide a description of the proposed method for one model consisting of one root filter and *n* parts.

**Input**: CP Decomposed filters set
Positive threshold set for pruning
Decomposition rank vector
Detection threshold $\tau$
**Result**: Set of detections D.
1  D←{empty set}
2  for $\gamma \in \mathfrak{I}$ do
3    for *i=0* to *n* do
4      p←0
5      for *r=1* to *R(n)*
6        p←p+$\varphi(\gamma) \otimes a_r \otimes b_r \otimes c_r$
7        if $p < t_r(i)$ then
8          s←-∞
9          skip $\gamma$
10       end
11     end
12     compute the best *i*-th part location,
        using deformations scores
13     compute an overall score s,
        using the best location for *i*-th part
14   end
15   if s ≥ $\tau$ then
16     D←D and {$\gamma$}
17   end
18 end
19 return D

Algorithm 1: Tensor based method with convolutions shortening.

## 5. Experiments

To test an overall speed-up gain and accuracy of the presented method we ran tests on a representative combined image set. Experiments were held for "person" and "car" categories. A compilation of INRIA and PASCAL VOC image sets was used for a person detection use case. A data set for a "car" category was constructed using PASCAL VOC, PBIC image libraries.

Resulting data sets appeared to be quite challenging for detection tasks due to significant lightning condition variations, large appearance of occlusions, deformations and poses. Authors consider such a combination to present higher representative ability. The cars data set consisted of more than 2500 images, and a person set was around 7000 image. Each data set was split into non-intersecting test and train parts.

The testing of proposed filters CP decomposition based method was performed on the basis of DPM release 5 [9]. One may check the proposed method on different types if detectors. For accuracy we compared misdetection rate versus false positive rate for various detector runs.

The method we proposed allows a natural speed-accuracy tradeoff. Bigger CP decomposition rank of the filter tends to higher accuracy of detection, but increases a number of convolutions to calculate the partial score. As a consequence, different decomposition ranks sets were tested.

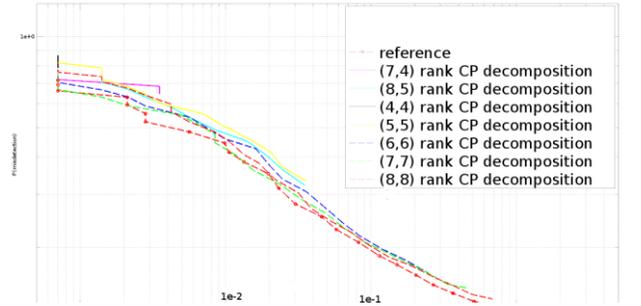

Figure 3: Root and part ranks CP decomposition impact. X axes is false positive, y-axes is misdetection frequency.

Here we perform *(S,T)* CP decomposition where root filters of person models were decomposed with rank "*S*" and all the rest part filters with rank "*T*".

An impact of filters CP decomposition ranks on accuracy is not strictly monotonic. E.g. (8,5) rank pair provides worse performance then (6,6), despite it has a better root filter approximation. The following figure 4 shows that a coarse approximation can provide near to reference accuracy: (6,11) CP decomposition has the same performance as (21,21) to within a small percentage of difference. A decomposition with rank pair (21,21) almost coincides with the reference curve.

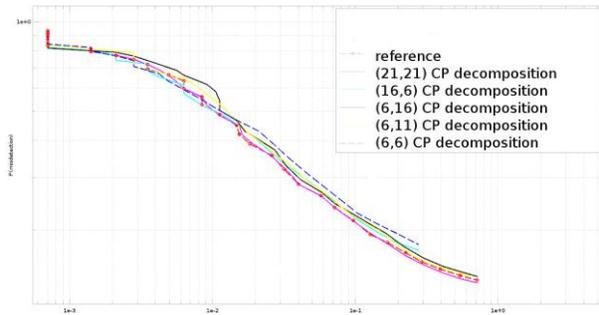

Figure 4: Root and part ranks CP decomposition impact. X axes stands for false positive frequency, y-axes is a misdetection rate.

According to formula (1), a speed-up gain of exploiting the proposed method with rank 6 for all filters and models is beyond 7 (for 8 part filters of size 8x8x32 and root size of 5x11x32). We succeeded in a search for a set of ranks such that the reference ROC-curve and the ROC-curve of tensor CP decomposition method match each other with accuracy to minimal threshold at every point in a region of interest. This rank set provides 4.5x speed-up compared to reference DPM v.5.

It can be noted that every filter can be stored in a decomposed form. Consequently we can replace storing the filter in memory as $NxMxL$ tensor with storing it as $R$ triples of vectors of length $N,M,L$. One should notice that a memory gain in this case also follows the formula (1).

## 6. Conclusion

Here we present a method which can improve DPM algorithm speed and memory usage. The method is based on a CP tensor decomposition of model filters and convolutions shortening using hypothesis pruning. Experiments show that this method provides up to 4.5 speed-up with no performance degradation and can provide a groundbreaking speed-up if some accuracy loss is allowable. Memory usage is also improved by a same multiplication factor. The method we proposed can be combined with other acceleration techniques (e.g. with sparselet or LUT based approaches) to make DPM truly applicable to real-time object detection scenarios.

This article provides next step towards solving speed bottleneck of deformable part models (DPM) algorithm or any other system that intensively uses convolutions operations (e.g. CNN)